# The RGNLP Machine Translation Systems for WAT 2018

**Atul Kr. Ojha**
SSIS, Jawaharlal Nehru University,
New Delhi, India
shashwatup9k@gmail.com

**Koel Dutta Chowdhury**
ADAPT Centre, Dublin City University
Dublin, Ireland
koel.chowdhury@adaptcentre.ie

**Chao-Hong Liu**
ADAPT Centre, Dublin City University
Dublin, Ireland
chaohong.liu@adaptcentre.ie

**Karan Saxena**
LTI, Carnegie Mellon University
Pittsburgh, PA, USA
karansax@cs.cmu.edu

## Abstract

This paper presents the system description of Machine Translation (MT) system(s) for Indic Languages Multilingual Task for the 2018 edition of the WAT Shared Task. In our experiments, we (the RGNLP team) explore both statistical and neural methods across all language pairs. (We further present an extensive comparison of language-related problems for both the approaches in the context of low-resourced settings.) Our PBSMT models were highest score on all automaticevaluation metrics in the English into Telugu, Hindi, Bengali, Tamil portion of the shared task.

## 1 Introduction

The Statistical Machine Translation (SMT) (Brown et al., 1993) has been a growing area in the Machine Translation (MT) for the last two decades in comparison to the Rule-based Machine Translation (RBMT), especially after the availability of Moses open source toolkit (Koehn et al., 2007). However, recent years have witnessed a surge in application of neural model for solving machine translation tasks. There are many NMT open source toolkits available such as OpenNMT (Klein et al., 2017), Neural Monkey (Helcl et al., 2017), Nematus (Sennrich et al., 2017) etc. With the goal of preventing low resource Indic languages from being left behind in the advancement of NMT, we take the first step towards applying neural methods for English⇆Indic Language pairs in the 2018 WAT Indic Languages Multilingual Task[1].

[1] http://lotus.kuee.kyoto-u.ac.jp/WAT/indic-multilingual/index.html

Our submission results show that despite being trained on the same training data, there are inconsistencies in translation quality between the SMT and NMT system. While NMT approaches continue to be a challenging problem in low-resource scenarios (Koehn et al., 2017), it clearly outperforms phrase based SMT model in terms of evaluation metrics for rich-resourced language pairs such as English-German, French-English, German-French, Russian-English, English-Czech, English-Chinese etc.

## 2 System Overview

We built 42 bidirectional MT systems (including 28 PBSMT and 14 NMT) for English⇄Indic language pairs. These were trained using both phrase-based statistical and neural network approaches. The system details are given below:

(a) **Phrase-based SMT Systems with KenLM and SRILM language model:** We built our phrase-based statistical MT systems using the Moses toolkit (Koehn et al., 2007). We use the GIZA++ (Och et al., 2003) toolkit with the grow-diag-final-and heuristic for extracting phrases from the corresponding parallel corpora. In addition, we use both KenLM and SRILM toolkits (Stolcke, 2002) to build 4-gram and 5-gram language models respectively. The KeNLM follows probing and TRIEs which renders the system to train faster (Heafield, 2011) while the SRILM follows TRIE (Stolcke, 2002). We use the scripts from Moses tokenizer to tokenize and lowercasing the English representations of our experiments.

(b) **Neural Machine Translation Systems on Long-Short Term Memory (LSTM) network:** To build our Neural Machine

Translation systems we use OpenNMT-py (the pytorch port of Open-NMT toolkit (Klein et al., 2017)). Our settings follow the Open-NMT training guidelines that indicate that the default training setup is reasonable for training any language pairs. Specifically, we use a 2-layer LSTM (Hochreiter et al, 1997) The model is trained for 13 epochs, using Adam (Kingma and Ba, 2015) with learning rate 0.002 and mini-batches of 40 with 500 hidden units, a vocabulary size of 50002 and 50004 respectively for the source and target-side of the data. We maintain a static NMT-setup using same hyper-parameters setting across all language pairs.

**(c) Direct Assessment and Ablation Study:** We evaluate our systems using three standard MT evaluation metrics- BLEU, RIBES, and AMFM scores. In addition to these, evaluation is also performed against direct Human evaluation metrics based on the JPOadequacy (Nakazawa, et al., 2016) for English and Hindi. 5 evaluators took part in the task over a period of approx.10 days to evaluate the translated outputs at sentence level. The final decisions were prepared by the means of voting. The scores were calculated and shared by WAT 2018 which have been shown and discussed in section 4 in detail.

## 3 Experiments

In this section, we briefly describe the experimental settings used to develop the PBSMT and NMT systems for seven Indic languages:

**Data Sets**
The data was provided by the WAT 2018 organizers under the **Indic Languages Multilingual Task**(Nakazawa et al., 2018). The parallel corpora were distributed as the '**Indic Languages Multilingual Parallel Corpus'**. These parallel corpora have been extracted from the Opus (OpenSubtitles) website which comes under the domain of spoken language. The detailed statistics of the parallel and monolingual corpora are demonstrated in Table-1 and 2 which used to train the MT systems. The parallel data was further divided into training, tuning and testing sets. The detailed information of the split is presented in Table-1.In terms of data volume, English⇌Singhalese language pair was the largest while English⇌Telugu language pair consists of minimum number of sentences. The similar trend is observed for the monolingual part of the corpora, with English having highestnumber of sentences and Telugu having the lowest.

| Language Pair | Training | Tuning | Testing | Total Parallel sentences (including training, tuning and testing ) |
|---|---|---|---|---|
| English⇌Hindi | 84557 | 500 | 1000 | 86057 |
| English⇌Bengali | 337428 | 500 | 1000 | 338928 |
| English⇌Malayalam | 359423 | 500 | 1000 | 360923 |
| English⇌Tamil | 26217 | 500 | 1000 | 27717 |
| English⇌Telugu | 22165 | 500 | 1000 | 23665 |
| English⇌Singhalese | 521726 | 500 | 1000 | 523226 |
| English⇌Urdu | 26619 | 500 | 1000 | 28119 |

Table 1: Statistics of Parallel Sentences of the Indic Multilingual Languages

| Language | Monolingual Sentences |
|---|---|
| English | 2891079 |
| Hindi | 104967 |
| Bengali | 453859 |
| Malayalam | 402761 |
| Tamil | 30268 |
| Telugu | 24750 |
| Singhalese | 705793 |
| Urdu | 29086 |

Table 2: Statistics of Monolingual Corpus of the Indic Multilingual Languages

### 3.1 Pre-Processing

For scope of this work, we perform the following Pre-processing steps. I Both types of corpora were tokenized, cleaned (removing sentences of length over 40 words). We also true-cased the English representations of the corpora. These processes were performed using Moses scripts. The tokenization of Indic languages was done by the RGNLP team tokenizer. The pre-processing of the Indic languages were done using tokenizer[2] provided by the RGNLP team to ensure the canonical Unicode representation.

### 3.2 Development of RGNLP Systems

In the next step, we developed three MT models perlanguage pair: two different phrase-based statistical machine translation system using

---
[2] https://github.com/shashwatup9k/

different language models and one neural MT system using the encoder-decoder framework.

### 3.2.1 Training and Developments of PBMST Systems

As above mentioned, we used the Moses open source tool the PBSMT system. The systems were trained independently and combined in a log-linear scheme in which each model was assigned a different weight using the Minimum Error Rate Training (Och et al., 2003) tuning algorithm. To investigate the role that language model has to play in terms of translation output, we used two different language model toolkits, namely KenLM and SRILM for building the 5-grams and 4-grams language models respectively. We used 500 parallel sentences for all language pairs to tune the systems.

### 3.2.2 Training and Developments of NMT Systems:

We use the OpenNMT toolkit for developing the NMT systems. We trained on a two layers of LSTM network with 500 hidden units at the both encoder and decoder models for 13 epochs. We have limited the variability of the parameters by using the default hyper-parameters configuration. Any unknown words in the translation were replaced with the word in the source language having the highest attention weight.

Finally, we translated the given test data using all 42 MT systems and performed some post-processing such as de-tokenization, de-truecasing to further improve the accuracy of the translated outputs.

## 4 Results and Analysis

In this section, we describe the following three things: (a) automatic evaluation results, (b) Human evaluation, and (c) Comparative Analysis of the PBSMT and NMT systems.

**(a) Automatic Evaluation Results:**

Evaluation is measured with the reference set provided the shared task organizers using the standard MT evaluation metrics. We present only the highest scoring system results across all language pair evaluated, in this paper. In order to gain a quantitative insight into specific differences, at least in terms of evaluation metrics, we highlight some results in Figure 1 and 2 as follows:

We see from the results that for PBSMT systems, the English-Hindi language pair produced best results in terms of all three metrics (44.08 in BLEU, 0.751 in RIBES, and 0.699 in AMFM) while the Malayalam-English language pair scored the lowest for all three metrics (8.74 BLEU).

For the NMT systems, the English⇌Hindi, English-Urdu scored the highest (21, 0.60, 0.47 in BLEU, RIBES and AMFM, respectively) while English-Singhalese scored 0.97 BLEU with respect to the SMT counter-part. Our PBSMT system highest and second highest scores with respect to BLEU and other evaluation metrics respectively across all language pair evaluated (shown in the Figure 3 and 4).

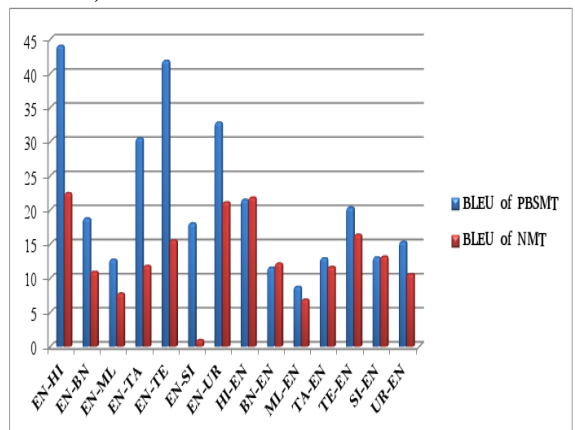

Figure 1: Accuracy of the English⇌Indic Languages of PBSMT and NMT Systems at the BLEU

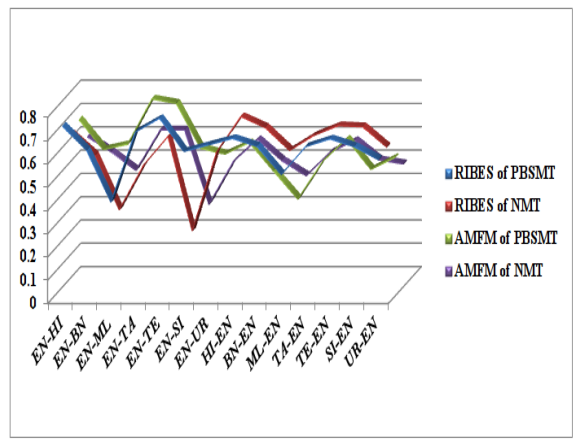

Figure 2: Accuracy of the English⇌Indic Languages of PBSMT& NMT Systems at the RIBES and AMFM

**(b) Human Evaluation Results:** In this section, we report the human evaluation accuracy of only English⇄HindiMT systems on adequacy. Figures 3 and 4 demonstrate the Pairwise and Adequacy results of English-Hindi and Hindi-English systems compared with other top MT systems. The Pairwise scores of our English-Hindi and Hindi-English systems were 15.50 and 22.25, respectively while the Adequacy of these pairs were 1.45 and 1.46. Both the Figures 3 and 4 clearly show that our systems hold the third rank in the human evaluation.

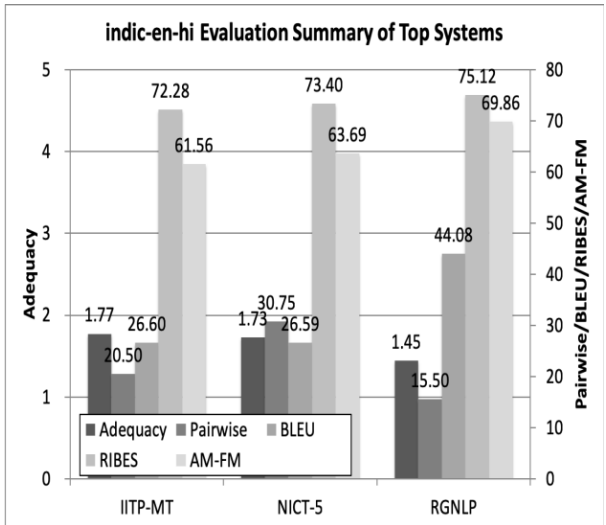

Figure 3: Comparative Evaluation of English-HindiMT Systems

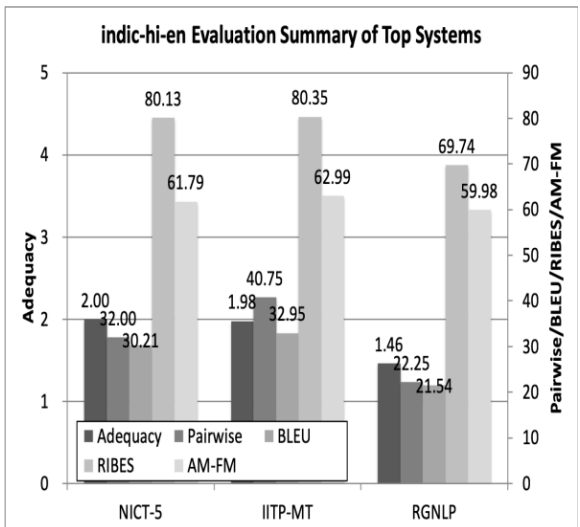

Figure 4: Comparative Evaluation of Hindi-English MT Systems

**(c) Comparative Analysis of the PBSMT and NMT Systems:** During comparison of the PBMST and NMT systems, the Indic-English language pairs of the NMT systems accuracies were the highest in BLEU, RIBES and AMFM metrics compared to other MT systems (Indic-English PBSMT, and English⇄Indic PBSMT and NMT), as shown in Figure 1 and 2. When we compare English-Hindi and Hindi-English both PBSMT and NMT systems at the adequacy level, the NMT's performance was worse (the accuracy was in negative). It happened because the NMT's result was affected majorly by over-generation, OOV (Out-of-Vocabulary), NER issues, and word-order and unable to produce output of some source sentences. The PBSMT's results were also affected by OOV, word-order, NER issues; nevertheless, it was able to produce output of each source sentence.

## 5   Conclusions

In this paper, two major points have been discussed. The first is development of the MT systems for English⇄Indic language pairs at the WAT2018 shared task and the second is the comparison of phrase-based statistical and neural based MT systems. The phrase-based and neural based MT systems were evaluated by automatic metrics on BLEU, RIBES and AMFM. To evaluate the adequacy of the PBSMT and NMT systems, the English-Hindi and Hindi-English MT systems were shared by five evaluators who evaluated these systems at the sentence level. The results of adequacy of systems were prepared via voting. Finally, we have compared and analyzed PBSMT and NMT systems and discussed their major problems.

**Acknowledgements**
We are grateful to the organizers of WAT2018 for providing us the Indic Language Multilingual Parallel and Monolingual Corpus and evaluation scores. We would also like to acknowledge the ADAPT Centre for Digital Content Technology which is funded under the SFI Research Centre Programme (Grant No. 13/RC/2106) and is co-funded under the European Regional Development Fund. This project has partially received funding from the European Union's Horizon 2020 Research and Innovation programme under the Marie Skłodowska-Curie Actions (Grant No. 734211).